\documentclass[11pt,a4paper]{article}
\usepackage[hyperref]{naaclhlt2019}
\usepackage{times}
\usepackage{latexsym}
\usepackage{graphicx}

\usepackage{amsmath}
\usepackage{amssymb}

\usepackage{tikz}
\usetikzlibrary{positioning}
\usetikzlibrary{arrows}

\aclfinalcopy 

\title{Aligning Very Small Parallel Corpora Using\\Cross-Lingual Word Embeddings and a Monogamy Objective}

\author{Nina Poerner, Masoud Jalili Sabet, Benjamin Roth and Hinrich Sch{\"u}tze \\
  Center for Information and Language Processing \\
  LMU Munich, Germany \\
  {\tt poerner@cis.uni-muenchen.de}}

\date{}

\begin{document}
\maketitle
\begin{abstract}
Count-based word alignment methods, such as the IBM models or fast-align, struggle on very small parallel corpora.
We therefore present an alternative approach based on cross-lingual word embeddings (CLWEs), which are trained on purely monolingual data.
Our main contribution is an unsupervised objective to adapt CLWEs to parallel corpora.
In experiments on between 25 and 500 sentences, our method outperforms fast-align. 
We also show that our fine-tuning objective consistently improves a CLWE-only baseline.
\end{abstract}

\section{Introduction}
Some parallel corpora, such as the Universal Declaration of Human Rights, are too small to apply count-based word alignment algorithms.

\newcite{sabet2016improving} show that integrating monolingual word embeddings into IBM Model 1 \cite{brown1990statistical} decreases word alignment error rate on a parallel corpus of 1000 sentences.
\newcite{pourdamghani2018using} exploit monolingual embedding similarity scores to create synthetic training data for Statistical Machine Translation (SMT), and report an increase in alignment F1. 

Recent advances have made it possible to create cross-lingual word embeddings (CLWEs) from purely monolingual data (\newcite{zhang2017adversarial}, \newcite{zhang2017earth}, \newcite{conneau2017word}, \newcite{artetxe2018robust}).
We propose to leverage such CLWEs for a \textbf{similarity-based} word alignment method, which works on corpora as small as 25 sentences.
Like \newcite{sabet2016improving}, our method relies only on monolingual data (to train the embeddings) and on the small parallel corpus itself.

Our \textbf{CLWE-only baseline} aligns source and target words in a parallel corpus if their CLWEs have maximum cosine similarity.
This approach is independent from the size of the parallel corpus, but has the following problems: 

\begin{itemize}
\item Semantics may differ between the embedding training domain and the parallel corpus.
\item CLWEs sometimes fail to discriminate between words with similar contexts, e.g., antonyms.
\end{itemize}

We therefore propose to \textbf{fine-tune} the CLWEs on the small parallel corpus using an \textbf{unsupervised embedding monogamy objective}.
To evaluate the proposed method, we simulate sparse data settings using Europarl sentences and Bible verses.
Our method outperforms the count-based fast-align model \cite{dyer2013simple} for corpus sizes up to 500 (resp., 250) sentences.
The proposed fine-tuning method improves over the CLWE-only baseline in terms of both precision and recall.

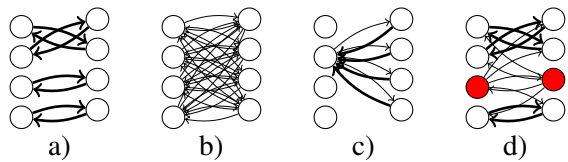
\begin{figure}[ht]
\centering
\begin{tikzpicture}[circ/.style={draw, circle, align=center, inner sep = 3pt},larr/.style={->, bend left = 15}, rarr/.style={->, bend left = 15}]
\foreach \i in {1,...,3} {
	\foreach \j in {0,...,3} {
		\node[circ] (l\i\j) at (2*\i, 0.4*\j) {};
		\node[circ] (r\i\j) at (2*\i+1, 0.4*\j+0.1) {};
	}
}

\foreach \j in {0,2,3} {
	\node[circ] (l4\j) at (8, 0.4*\j) {};
	\node[circ] (r4\j) at (9, 0.4*\j+0.1) {};
}

\node[circ, fill=red] (l41) at (8, 0.4) {};
\node[circ, fill=red] (r41) at (9, 0.5) {};

\node [align=center] at (2.5, -0.4) {a)};
\node [align=center] at (4.5, -0.4) {b)};
\node [align=center] at (6.5, -0.4) {c)};
\node [align=center] at (8.5, -0.4) {d)};

\foreach \j in {0,...,3} {
	\foreach \k in {0,...,3} {
		\path(l2\j) edge[larr, line width = .25] (r2\k);
		\path(r2\k) edge[rarr, line width = .25] (l2\j);
	}
	\path(r3\j) edge[rarr, line width = 1] (l32);
	\path(l32) edge[larr, line width = .25] (r3\j);
}

\foreach \j in {0,1} {
	\path(l1\j) edge[larr, line width = 1] (r1\j);
	\path(r1\j) edge[rarr, line width = 1] (l1\j);
}

\path(l13) edge[larr, line width = 1] (r12);
\path(r12) edge[rarr, line width = 1] (l13);
\path(l12) edge[larr, line width = 1] (r13);
\path(r13) edge[rarr, line width = 1] (l12);

\path(l40) edge [larr, line width = 1] (r40);
\path(r40) edge [rarr, line width = 1] (l40);

\foreach \j in {0,...,3} {
	\path(l41) edge [larr, line width = .25] (r4\j);
	\path(r41) edge [rarr, line width = .25] (l4\j);
}

\path(l42) edge [larr, line width = 1] (r43);
\path(r42) edge [rarr, line width = 1] (l43);
\path(l43) edge [larr, line width = 1] (r42);
\path(r43) edge [rarr, line width = 1] (l42);

\end{tikzpicture}
\caption{Schematic representation of the monogamy objective. a) one-to-one (``monogamous'') alignment: $l(s,t) = 0$, b) many-to-many alignment: $l(s,t) = 1$, c) one-to-many alignment: $l(s,t) = 1$, d) minimizing $l(s,t)$ means making the red nodes more similar to each other, and less similar to the white nodes.}
\label{fig:motivation}
\end{figure}
 
\section{Method}
\subsection{CLWE-only baseline}
\label{sec:baseline}
Our CLWE-only baseline uses a cross-lingual embedding space derived from purely monolingual data \cite{artetxe2018robust}.
Let $D$ be our small corpus, and let $s$ (source) and $t$ (target) be parallel sentences from $D$.
Let $\mathrm{clwe}(s_i)$ and $\mathrm{clwe}(t_j)$ be the embedding vectors of tokens $s_i$ and $t_j$.
We align $s_i$ to $\mathrm{argmax}_{t_j \in t} [\mathrm{cos}(\mathrm{clwe}(s_i), \mathrm{clwe}(t_j)]$.
Any ties are broken by proximity to the diagonal of the alignment matrix.

\subsection{Fine-tuning method}
\paragraph{Intuition.}
Assume that we have the following sentence pair: \textit{aaa bbb xxx} $\lvert\lvert\lvert$ \textit{111 000 222}.
Assume further that we know from CLWEs that \textit{aaa} $\approx$ \textit{111} and \textit{bbb} $\approx$ \textit{222}, but we lack informative embeddings for \textit{000} and \textit{xxx}.
We may hypothesize that $\textit{xxx} \approx \textit{000}$, as they are the only tokens that lack translations.
We may also hypothesize that $\textit{xxx} \not \approx \textit{111}$, $\textit{xxx} \not \approx \textit{222}$, as $111$ and $222$ already have translations of their own.

In the following, we will refer to this principle as \textbf{embedding monogamy}. 
We assume that in the absence of evidence to the contrary, a source embedding should have
\begin{itemize}
\item high similarity to one target embedding
\item low similarity to other target embeddings\footnote{
Of course, this assumption is over-simplistic, as one-to-n alignments exist (e.g., English \textit{not} should be similar to both French \textit{ne} and \textit{pas}).}
\end{itemize}
This principle is related to the IBM Model \cite{brown1990statistical}, where Expectation Maximization increases $p(f|e)$ if $e$ and $f$ co-occur in sentences where $f$ is not explained by other source words.

\paragraph {Embedding monogamy objective.}
We define the probability of $t_j$ given $s_i$ as:

\begin{equation}
\displaystyle
\label{eq:prob}
p(t_j | s_i, t) = \frac{e^{\frac{1}{\tau} \mathrm{cos}(\mathrm{clwe}(s_i), \mathrm{clwe}(t_j))}}{\sum_{j'} e^{\frac{1}{\tau} \mathrm{cos}(\mathrm{clwe}(s_i), \mathrm{clwe}(t_{j'}))}}
\end{equation}
where $\tau$ is a temperature hyperparameter.
This definition is similar to the definition of translation probability in \newcite{artetxe2018unsupervised} and \newcite{lample2018phrase}.
But while they normalize over the vocabulary, we normalize over the target sentence.
As a consequence, the probability of $t_j$ depends not only on $s_i$, but also on competitor tokens in $t$.

With these translation probabilities, we model a two-step random walker $\mathbf{R}^{s \rightarrow t \rightarrow s}$ that starts at $s_i$, steps to a random target word and then to $s_{i'}$: $r^{s \rightarrow t \rightarrow s}_{ii'} = \sum_{j=1}^{\mathrm{len}(t)} p(t_j | s_i, t) p(s_{i'} | t_j, s)$.
To maximize monogamy, we maximize the entries on the diagonal of $\mathbf{R}^{s \rightarrow t \rightarrow s}$, i.e., the probability of the walker returning to its origin.
To avoid penalizing long sentences, we minimize the negative logarithm to the base of the source sentence length: $l(s, t) = 1 -\mathrm{log}_{\mathrm{len}(s)} \sum_{i=1}^{\mathrm{len}(s)} r^{s \rightarrow t \rightarrow s}_{ii}$. 
This loss has the following properties:
\begin{itemize}
\item In a fully ``monogamous'' situation (see Figure \ref{fig:motivation} a), $r^{s \rightarrow t \rightarrow s}_{ii} \rightarrow 1 \implies l(s,t) \rightarrow 0$.
\item In a situation where all source words are equidistant from all target words (see Figure \ref{fig:motivation} b), $r^{s \rightarrow t \rightarrow s}_{ii} = \frac{1}{\mathrm{len}(s)} \implies l(s,t) = 1$.
\end{itemize}

Reversing the roles of source and target results in the following bidirectional loss: $L_\mathrm{bi}(s,t) = \frac{1}{2} [l(s,t) + l(t,s)]$.
Both terms are necessary, since a given alignment may appear highly monogamous from the perspective of one sentence but not the other (especially when there are left-over words due to a difference in length).

\paragraph{Adding position information.}
At this point, our objective ignores word positions, which we know to be useful from count-based methods (e.g., \newcite{dyer2013simple}).
Therefore, we add position embeddings inside the translation probability equation:

\begin{equation}
\displaystyle
\nonumber
p(t_j | s_i, t) = \frac{e^{\frac{1}{\tau} \mathrm{cos}[\mathrm{clwe}(s_i) + \mathrm{a}(i), \mathrm{clwe}(t_j) + \mathrm{a}(j)]}}{\sum_{j'} e^{\frac{1}{\tau} \mathrm{cos}[\mathrm{clwe}(s_i) + \mathrm{a}(i), \mathrm{clwe}(t_{j'}) + \mathrm{a}(j')]}}
\end{equation}
where $\mathrm{a}(i)$ is a sinusoid embedding vector for position $i$ \cite{vaswani2017attention}.
As a result, word pairs near the diagonal have higher round trip probabilities initially.
Since the monogamy objective aims to strengthen strong links, similar position embeddings act as attractors for non-positional embeddings.
Note that we use only the non-positional embeddings for alignment, as the position prior is too strong at test time.

\paragraph{Alignment retention objective.}
In initial experiments, we found that the monogamy objective increases recall but risks losing precision, relative to the CLWE-only baseline.
Therefore, we add an additional objective that aims to increase round trip probability for alignments made by the baseline, but does not influence unaligned words:

\begin{equation}
\displaystyle
\begin{aligned}
\nonumber
L_\mathrm{ret}(s,t) & = \frac{1}{2} [l_\mathrm{ret}(s,t) + l_\mathrm{ret}(t,s)] \\ \nonumber
l_\mathrm{ret}(s,t) & = - \mathrm{log} \frac{\sum_{i,j} p(t_j | s_i, t) p(s_i | t_j, s) m^{st}_{ij}}{\sum_{i,j} m^{st}_{ij}} \\ \nonumber
m^{st}_{ij} & = \mathbb{I}[(s_i, t_j) \in \mathrm{align}_0] \\ \nonumber
\end{aligned}
\end{equation}
where $\mathrm{align}_0$ is the intersection of the $s$-to-$t$ and $t$-to-$s$ alignments made with the initial CLWEs (see Section \ref{sec:baseline}). 
Our final loss function is: $L(D) = \frac{1}{|D|} \sum_{(s,t) \in D} [L_\mathrm{bi}(s,t) + \alpha L_\mathrm{ret}(s,t)]$.

\begin{figure}[t]
\centering
\includegraphics[scale=.63]{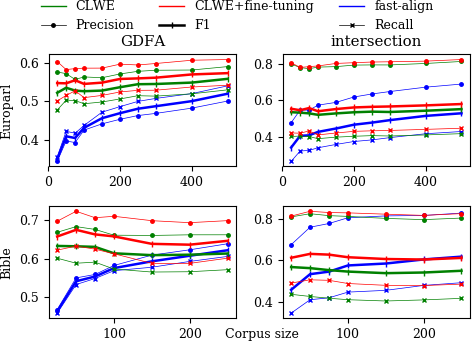}
\caption{Alignment precision, recall and F1 as a function of corpus size.}
\label{fig:results}
\end{figure}

\section{Evaluation}
\label{sec:eval}
We evaluate our model on subsets of different sizes from the English-German Europarl gold alignments\footnote{\url{www-i6.informatik.rwth-aachen.de/goldAlignment/}} and French-English Bible gold alignments \cite{melamed1998manual}\footnote{\url{nlp.cs.nyu.edu/blinker/}. We consider links with inter-annotator agreement as sure, others as possible.}.
We initialize CLWEs with the unsupervised algorithm of \newcite{artetxe2018robust} on monolingual FastText embeddings \cite{bojanowski2017enriching}\footnote{\url{fasttext.cc}, top-200000 words per language}.
Fine-tuning is done in \texttt{keras}, using the adam optimizer \cite{kingma2014adam}.
We set $\alpha = 1.0$ and $\tau = 0.001$, and apply $50\%$ dropout to the embeddings.

We use fast-align \cite{dyer2013simple} as a count-based baseline, since it outperformed the IBM models in initial experiments.
We symmetrize alignments by either intersection or the grow-diag-final-and (GDFA) heuristic \cite{koehn2007moses}.
We train fast-align and our fine-tuning method for 500 iterations.

\section{Discussion}
\subsection{Corpus size}
The performance of fast-align is highly dependent on corpus size, which is not surprising, seeing that it has to infer word semantics from the small corpus alone.
The CLWE-only baseline on the other hand is independent from corpus size, resulting in decent performance even on 25 parallel sentences.
Importantly, the positive effect of our fine-tuning method seems to be robust to corpus size, as we see improvements in F1 for all sizes.

\subsection{Benefits of fine-tuning}
We find that the proposed fine-tuning method has a positive effect on alignment precision and recall, relative to the CLWE-only baseline.
We assess some sentence pairs qualitatively to find reasons for this improvement:

\begin{figure*}[t]
\centering
\includegraphics[scale=.72]{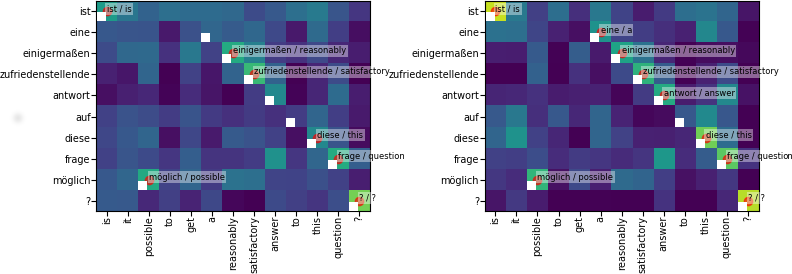}
\includegraphics[scale=.72]{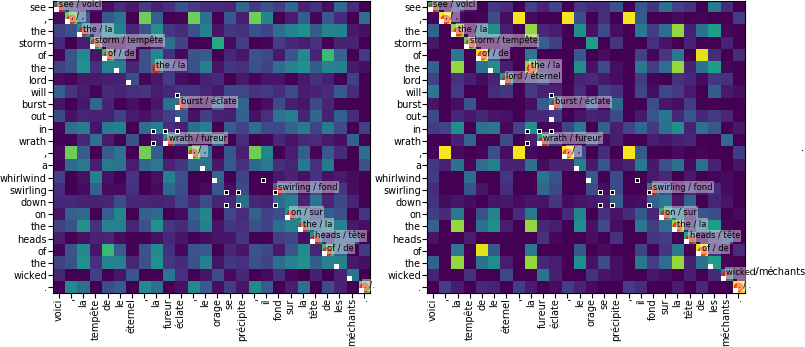}
\caption{Similarity matrices before (left) and after (right) fine-tuning. Red dots: our alignment (intersection). White squares: sure gold alignments. Empty white squares: possible gold alignments.}
\label{fig:ex}
\end{figure*}

\paragraph{Resolution of ambiguities.}
Word embeddings sometimes fail to differentiate between words with similar contexts, such as antonyms. 
In Figure \ref{fig:ex} (top), our fine-tuning method resolves such an ambiguity: 
Here, the initial CLWE of \textit{answer} is slightly more similar to German \textit{frage} (= \textit{question}) than to the true translation \textit{antwort}.
Since \textit{frage} already has a round trip partner, the monogamy objective pushes \textit{answer} away from \textit{frage}, resulting in the addition of a correct alignment between \textit{answer} and \textit{antwort}.

\paragraph{In-domain word translations.}
Since word embeddings are trained on general-purpose corpora, CLWEs can fail to reflect domain-specific word translations. 
One such example is the translation of \textit{lord} as French \textit{{\'e}ternel} ($\approx$ \textit{``eternal one''}) in Figure \ref{fig:ex} (bottom).
While the translation is common in this particular Bible version, the CLWEs do not reflect it well ($\mathrm{cos}(\textit{lord}, \textit{{\'e}ternel}) < \mathrm{cos}(\textit{wicked}, \textit{{\'e}ternel})$).
Through fine-tuning, and due to their frequent coocurrence in the small corpus, the similarity between \textit{{\'e}ternel} and \textit{lord} increases enough for a successful alignment.

\section{Use case: Aligning the UDHR}
\begin{figure*}
\centering
\includegraphics[scale=.72]{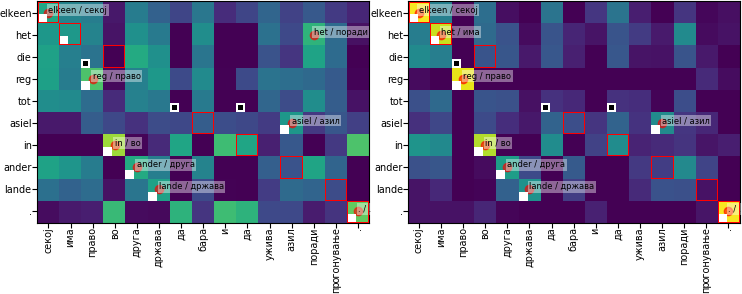}
\includegraphics[scale=.72]{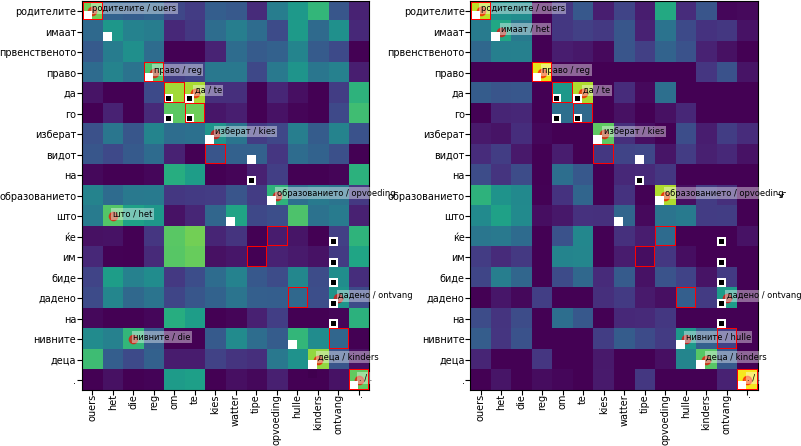}
\caption{Articles 14(1) and 26(3) from the UDHR. Similarity matrices before (left) and after (right) fine-tuning. Red dots: our alignment (intersection). Red boxes: fast-align (intersection). White squares: sure gold alignments. Empty white squares: possible gold alignments (by the authors).}
\label{fig:mk-af}
\end{figure*}

In practice, our method would not be applied to English-German or English-French, as there is no lack of parallel data for these language pairs.
For a more realistic use case, we align the 50 articles of the Universal Declaration of Human Rights\footnote{https://unicode.org/udhr/} in Macedonian and Afrikaans.
While we do not have gold alignments for an evaluation, a preliminary qualitative analysis suggests that our method finds a reasonable semantic word alignment, while fast-align mainly predicts the diagonal (see Figure \ref{fig:mk-af} for examples).

\section{Related Work}
\paragraph{Embeddings for word alignment.}
\newcite{sabet2016improving} reformulate the IBM 1 model to predict the probability of monolingual target embedding vectors.
They report improvements in AER for English-French on parallel corpora between 1K and 40K sentences, as well as improvements in precision on words with frequency $\leq$ 20.

\newcite{pourdamghani2018using} exploit similarity scores from monolingual embeddings to create synthetic training data for an SMT system.
They report improvements for English-Chinese, English-Arabic and English-Farsi alignment ($\Delta F1 = 0.2\%, 0.5\%, 1.7\%$). 
Their smallest parallel corpus has 500K sentences, while we align a few dozen to hundred sentences.

\paragraph{Two-step round trip objective.}
Our use of two-step round trips is inspired by \newcite{haeusser2017associative}. They optimize domain adaptation using a random walker that steps from image representations with known labels to image representations with unknown labels and back.
While their target is a uniform distribution over images with the same label as the image of origin, ours is to have maximum probability mass on the word of origin.

\paragraph{Low resource CLWEs.}
Our approach relies on the availability of high-quality CLWEs.
\newcite{wada2018unsupervised} report that in settings with little monolingual data ($<$ 1M sentences), mapping approaches like \newcite{artetxe2018robust} are not robust.
Instead, they propose to learn CLWEs from a language model trained on the union of two small monolingual corpora.
Their work is orthogonal to our fine-tuning method, since we make no assumptions about how the CLWEs are created.

\section{Conclusion}
We have presented a \textbf{similarity-based} method to produce word alignments for very small parallel corpora, using monolingual data and the corpus itself.
Our \textbf{CLWE-only baseline} uses an unsupervised mapping of monolingual embeddings \cite{artetxe2018robust}.
Our main contribution is an \textbf{unsupervised embedding monogamy objective}, which adapts CLWEs to the small parallel corpus.
Our model outperforms count-based fast-align \cite{dyer2013simple} on parallel corpora up to 500 (resp., 250) sentences.

We expect that our method will be useful in low-resource settings, e.g., when aligning the Universal Declaration of Human Rights.

\paragraph{Acknowledgments.}
We gratefully acknowledge funding for this work by the European Research Council (ERC \#740516).

\bibliographystyle{acl_natbib}
\bibliography{main}

\end{document}